\documentclass[10pt,twocolumn,letterpaper]{article}

\usepackage{iccv}
\usepackage{times}
\usepackage{epsfig}
\usepackage{graphicx}
\usepackage{amsmath}
\usepackage{amssymb}
\usepackage{xcolor}
\usepackage{multirow}

\usepackage[breaklinks=true,bookmarks=false,citecolor=green]{hyperref}

\iccvfinalcopy 

\def\iccvPaperID{****} 

\ificcvfinal\fi

\begin{document}

\title{Learning Sparse 2D Temporal Adjacent Networks \\for Temporal Action Localization}


\author{Songyang Zhang$^{1,2}$
\thanks{This work was partially done in Microsoft Research}
\quad Houwen Peng$^2$ \quad Le Yang$^2$ \quad Jianlong Fu$^2$ \quad Jiebo Luo$^1$\\
$^1$University of Rochester \qquad\qquad $^2$Microsoft Research\\
{\tt\small szhang83@ur.rochester.edu,\{houwen.peng,v-yale,jianf\}@microsoft.com,jluo@cs.rochester.edu}}

\maketitle
\ificcvfinal\thispagestyle{empty}\fi

\begin{abstract}

In this report, we introduce the Winner method for HACS Temporal Action Localization Challenge 2019.
Temporal action localization is challenging since a target proposal may be related to several other candidate proposals in an untrimmed video.
Existing methods cannot tackle this challenge well since temporal proposals are considered individually and their temporal dependencies are neglected. To address this issue, we propose sparse 2D temporal adjacent networks to model the temporal relationship between candidate proposals. This method is built upon the recent proposed 2D-TAN approach~\cite{2DTAN_2020_AAAI}. The sampling strategy in 2D-TAN introduces the unbalanced context problem, where short proposals can perceive more context than long proposals.
Therefore, we further propose a Sparse 2D Temporal Adjacent Network (S-2D-TAN).
It is capable of involving more context information for long proposals and further learning discriminative features from them. By combining our S-2D-TAN with a simple action classifier, our method achieves a mAP of $23.49$ on the test set, which win the first place in the HACS challenge.

\end{abstract}

\section{Methodology}
\begin{figure}[t!]
\centering
\includegraphics[width=0.5\textwidth]{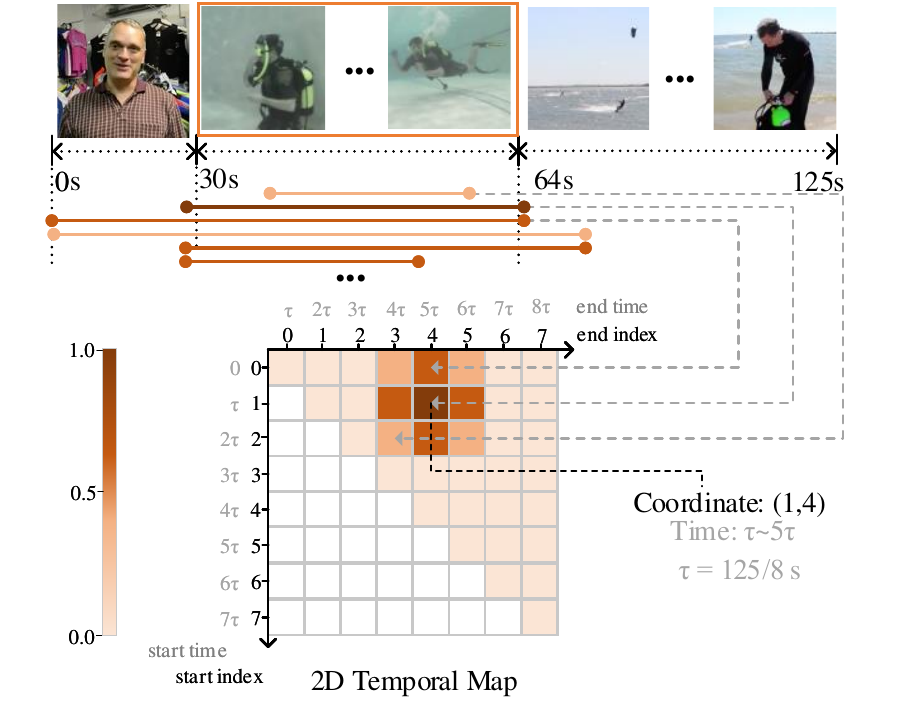}
\caption{Examples of localizing actions in an untrimmed video. 
In the two-dimensional temporal map, the \textit{black} vertical and horizontal axes represent the start and end frame indices while the corresponding \textcolor{gray}{\textit{gray}} axes represent the corresponding start and end time in the video.
The values in the 2D map, highlighted by red color, indicate the overlapping scores between the candidate proposals and the target proposal.
Here, $\tau$ is a short duration determined by the video length and sampling rate.
}
\label{fig:task}
\end{figure}

\begin{figure*}[!thb]
\centering
\includegraphics[width=\textwidth]{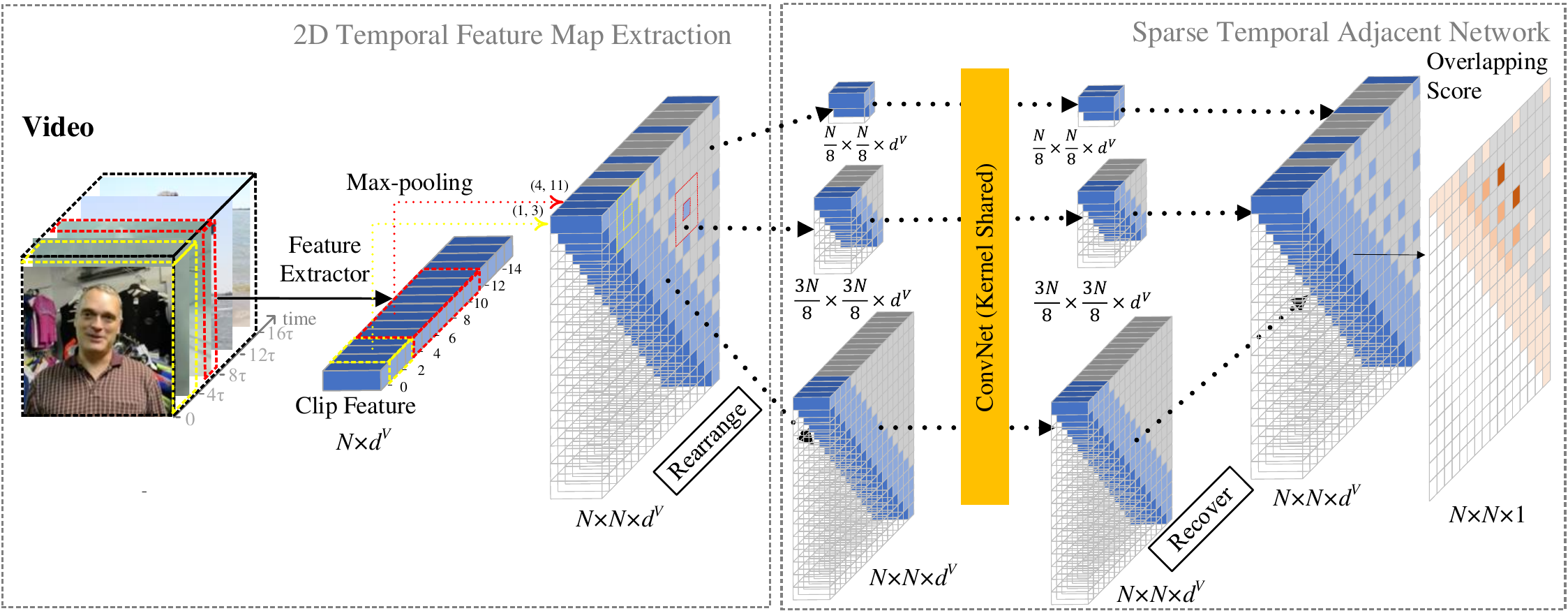}
\caption{The framework of our proposed Sparse 2D Temporal Adjacent Network. It consists of a 2D temporal feature map extractor for video representation and a temporal adjacent network for proposal generation.
\textcolor{blue}{\textit{Blue}} boxes represent the proposals we select and the \textcolor{gray}{\textit{gray}} boxes represent the proposals that are not selected. All the \textit{transparent} boxes represent the invalid proposals.
\textcolor{blue}{\textit{Blue}} boxes with \textcolor{violet}{\textit{violet }}, \textcolor{orange}{\textit{orange}} and \textcolor{green}{\textit{green}} margins represent short, medium and long proposals we selected.
}
\label{fig:framework}
\end{figure*}

Our framework is inspired from the concept of 2D Temporal map~\cite{2DTAN_2020_AAAI}, which is originally designed for the moment localization with natural language task. 
We extend this concept to the temporal action localization task.
The core idea is to design a 2D temporal map, where  one dimension  indicates  the  starting  time  of  a  proposal and the other indicates the end time, as shown in Figure~\ref{fig:task}.
In the following, we introduce our proposed Sparse 2D Temporal Adjacent Network approach. This approach consists of four steps: video representation, proposal generation, action classification, and score fusion.
Figure~\ref{fig:framework} shows the framework of the proposed S-2D-TAN approach.

\subsection{Video Representation via 2D Temporal Feature Map}
\label{sec:2D-Map}
In this section, we extracts the features from the input video stream via slowfast network~\cite{feichtenhofer2019slowfast} and then encodes the features to a 2D Temporal Feature Map (see 2D-TAN~\cite{2DTAN_2020_AAAI} for more details.).
Since extracting features for all possible proposals on the map is time consuming, we also follow the same sampling strategy in 2D-TAN~\cite{2DTAN_2020_AAAI}, where short proposals are densely selected and long proposals are sparsely selected as candidates.
In order to reduce the redundant max operations and number of parameters, we stack max-pooling layers to extract proposal features, similar to the stacked convolutions in previous work~\cite{zhang2018s3d,lin2017single}.

\label{sec:S-2D-TAN}
\subsection{Proposal Generation via Sparse 2D Temporal Adjacent Network}

In this section, we predict the possibility of candidate proposals overlapping with the target proposals.
The original 2D-TAN network is simple and only consists of several convolution layers.
When the convolution is operated, invalid proposals are not involved in the computation.
However, this design ignore the effects of different number of context proposals: short proposals perceive more context proposals  (\textit{e.g.} \textcolor{red}{\textit{red}} dashed box) compared to long proposals (\textit{e.g.} \textcolor{yellow}{\textit{yellow}} dashed box), as shown in Figure~\ref{fig:framework}.
In contrast, we rearrange the original sparse feature map into three compact sub feature maps and separately feed them to a shared convolution network.
More details are shown as following:

Let a proposal start from $i$-th clip to $j$-th clip and its length $l_{ij}=j-i+1$.
\begin{itemize}
    \item For short proposals that satisify $l_{ij} \le \frac{N}{4}$, we enumerate all possible proposals as candidates and rearrange them to a new feature map of size $N\times N$.
    \item For medium proposal that satsify $\frac{N}{4} < l_{ij}\le \frac{N}{2}$, we select them with stride $2$ and rearrange them into a compact map of size $\frac{3N}{8}\times \frac{3N}{8}$.
    \item For long proposal that satsify $\frac{N}{2} < l_{ij}\le N$, we select them with stride $4$ and rearrange them into a compact map of size $\frac{N}{8}\times \frac{N}{8}$.
\end{itemize}

These three compact feature maps are then separately fed into a shared convolution network.
For each map, we keep the input and output sizes same by using zero padding.
The outputs are then recovered to a feature map. The recovered locations are based on its corresponding location in the original map before the rearrangement.

Finally, we predict the overlapping scores of proposals on the 2D temporal map. The output feature of sparse temporal adjacent network are passed through a fully connected layer and a sigmoid function. The output value $\mathbf{P_{over}}\in \mathbb{R}^{N\times N\times 1}$ indicates the possibility of the proposal contains pre-defined actions.

For the loss function, we adapt a soft binary cross entropy loss~\cite{2DTAN_2020_AAAI} with two $IoU$ thresholds $t_{min}$ and $t_{max}$ for training. 
Noted that only valid proposals on the map are involved during loss computation and inference.

\subsection{Action Classifier}

In this section, we independently train an action classifier which predict the class label from a given proposal.
In more details, we select proposals that have $IoU$ larger than $0.5$ with ground truth as training samples and mark all the other as invalid.
Their proposal features are obtained same as section~\ref{sec:2D-Map}.
Without any convolution layers, the proposal features goes through a fully connected layer and a softmax layer to obtain the classification scores $\mathbf{P_{class}}\in \mathbb{R}^{N\times N\times C}$, where $C$ is the total number of classes.
If there are multiple actions involved in one proposal, we choose the action with highest $IoU$ as its label.
Cross-entropy loss is used for training.

\begin{table*}[thb!]
\begin{center}
\begin{tabular}{ |c|c|c|c|c|c|c|c|} 
 \hline
 \multirow{3}*{Row\#} & \multirow{3}*{Model} & \multicolumn{3}{c|}{Parameters} & \multirow{3}*{Number of Layers} & \multirow{3}*{AR@$100$} & \multirow{3}*{AUC}\\
 \cline{3-5}
 & & Number of  & Number of  & \multirow{2}*{Kernel Size} &  &  & \\ 
 & & Max-Pooling Layers & Clips &  & & & \\ 
 \hline
 1 & S-2D-TAN & 128 & 256 & 9 & 4 & \textbf{67.25} & \textbf{62.40} \\ 
 2 & 2D-TAN & 128 & 128 & 9 & 4 & 63.25 & 58.11 \\ 
 3 & 2D-TAN & 64 & 64 & 9 & 4 & 55.24 & 50.40 \\ 
 4 & 2D-TAN & 64 & 64 & 1 & 1 & 49.85 & 42.32 \\ 
 \hline
\end{tabular}
\end{center}
\caption{Ablation Study for the proposal generation. Experiments are conducted on the HACS validation set. AR@$100$ and AUC follows the same definition in ActivityNet~\cite{caba2015activitynet}.}
\end{table*}

\subsection{Score Fusion}

In this step, we predict the final detection results.
Following Lin \textit{et al.}~\cite{lin2017single}, the final score is computed by multiplying the overlapping score and classification score, as shown in the following: 
\begin{equation}
    \mathbf{P_{final}} = \mathbf{P_{over}}\mathbf{P_{class}}\in \mathbb{R}^{N\times N\times C}
\end{equation}
We then apply Non-maximum-suppression (NMS) to $\mathbf{P_{final}}$ and obtain the final predictions.

\section{Experiments}

We evaluate our proposed method on the HACS validation set~\cite{zhao2019hacs}. 
In this section, we first introduce our implementation details and then investigate the impact of different factors through a set of ablation studies.

\subsection{Implementation Details}
Both the S-2D-TAN network and the classifier are optimized by Adam~\cite{kingma2014adam} with learning rate of $1\times 10^{-3}$ and batch size of $32$. 
The size of all hidden states in the model is set to $512$.
For S-2D-TAN network, a $4$-layer convolution network with kernel size of $9$ is adopted, and non maximum suppression (NMS) with a threshold of $0.5$ is applied during the inference.
The scaling thresholds $t_{min}$ and $t_{max}$ are set to $0.5$ and $1.0$.
During our experiments, the number of sampled clips is set to $256$, which means the map’s spatial dimension is of size $256\times256$. 
We densely select all short proposals as candidates if the length is less than $64$. 
For proposals with medium length between $64$ and $128$, the sampling stride between candidates are increased by $2$. 
For long proposals with length above $128$, the sampling stride is further increased to $4$.
The network has $64+64/2+128/4=128$ stacked max-pooling layers in total.

\subsection{Experiment Results}

In this section, we evaluate the effects of different factors. 
We can observe that by enlarging the receptive field, the performance increase significantly (Row $3$ \textit{v.s.} Row $4$).
If we keep the receptive field same and increasing the number of proposal candidates, the performance can get further improvement (Row $2$ \textit{v.s.} Row $3$).
After adopting our proposed sparse convolution strategy, the performance is further improved (Row $1$ \textit{v.s.} Row $2$).
By combining the best proposal generation model (Row $1$) with a separately trained classifier, we can achieve $23.49$ mAP in the HACS test set.

\section{Conclusion}

In this report, we extends the 2D-TAN approach to the temporal action localization task. We also introduce a novel Sparse 2D Temporal Adjacent Network (S-2D-TAN) to handle the unbalanced context proposals. The performance on HACS dataset has verify its effectiveness. In the future, we will conduct more experiments on other datasets, like ActivityNet, THUMOS to test our model's performance.

{\small
\bibliographystyle{ieee_fullname}
\bibliography{reference}
}

\end{document}